\documentclass[12pt]{report}

\usepackage{tamuconfig}

\usepackage{times}
\usepackage[T1]{fontenc}

\usepackage{graphicx}

\DeclareGraphicsExtensions{.png}

\graphicspath{ {./graphic/} }

\usepackage[hidelinks]{hyperref}

\begin{document}

\renewcommand{\tamumanuscripttitle}{Memory Access Characterization of Large Language Models in CPU Environment and Its Potential Impacts}

\renewcommand{\tamupapertype}{Thesis}

\renewcommand{\tamufullname}{Spencer Matthew Banasik }

\renewcommand{\tamudegree}{Master of Science}
\renewcommand{\tamuchairone}{Abdullah Muzahid}

\renewcommand{\tamumemberone}{Paul Gratz}
\newcommand{\tamumembertwo}{Chia-Che Tsai}
\newcommand{\tamumemberthree}{Committee Member 3}
\renewcommand{\tamudepthead}{Scott Schaefer}

\renewcommand{\tamugradmonth}{May}
\renewcommand{\tamugradyear}{2025}
\renewcommand{\tamudepartment}{Computer Science}

%
%
%
%
%


\providecommand{\tabularnewline}{\\}

\begin{titlepage}
\begin{center}
\begin{doublespace}

\MakeUppercase{  \tamumanuscripttitle}
\end{doublespace}
\vspace{4em}

A \tamupapertype

by

\MakeUppercase{\tamufullname}

\vspace{4em}

\begin{singlespace}

Submitted to the Graduate and Professional School of \\
Texas A\&M University \\

in partial fulfillment of the requirements for the degree of \\
\end{singlespace}

\MakeUppercase{\tamudegree}
\par\end{center}
\vspace{2em}

\begin{singlespace}
\begin{tabular}{ll}
 & \tabularnewline
& \cr
Chair of Committee, & \tamuchairone\tabularnewline
Committee Members, & \tamumemberone\tabularnewline
 & \tamumembertwo\tabularnewline
Head of Department, & \tamudepthead\tabularnewline

\end{tabular}
\end{singlespace}
\vspace{3em}

\begin{center}
\tamugradmonth \hspace{2pt} \tamugradyear

\vspace{3em}

Major Subject: \tamudepartment \par
\vspace{3em}
Copyright \tamugradyear \hspace{.5em}\tamufullname 
\par\end{center}
\end{titlepage}
\pagebreak{}


%
%
%
%
%

\chapter*{ABSTRACT}
\addcontentsline{toc}{chapter}{ABSTRACT} 

\pagestyle{plain} 
\pagenumbering{roman} 
\setcounter{page}{2}

\indent As machine learning algorithms are shown to be an increasingly valuable tool, the demand for their access has grown accordingly. Oftentimes, it is infeasible to run inference with larger models without an accelerator, which may be unavailable in environments that have constraints such as energy consumption, security, or cost. To increase the availability of these models, we aim to improve the LLM inference speed on a CPU-only environment by modifying the cache architecture. To determine what improvements could be made, we conducted two experiments using Llama.cpp and the QWEN model: running various cache configurations and evaluating their performance, and outputting a trace of the memory footprint. Using these experiments, we investigate the memory access patterns and performance characteristics to identify potential optimizations.

\pagebreak{}

%
%
%
%
%

\chapter*{DEDICATION}
\addcontentsline{toc}{chapter}{DEDICATION}  

\begin{center}
\vspace*{\fill}
To my mother, father, and sister. 
\vspace*{\fill}
\end{center}

\pagebreak{}

%
%
%
%
%

\chapter*{CONTRIBUTORS AND FUNDING SOURCES}
\addcontentsline{toc}{chapter}{CONTRIBUTORS AND FUNDING SOURCES}  

\subsection*{Contributors}
This work was supported by a thesis committee consisting of Professor Abdullah Muzahid and Chai-Che Tsai of the Department of Computer Science and Engineering and Professor Paul Gratz of the Department of Electrical and Computer Engineering.


All other work conducted for the thesis was completed by the student independently.

\subsection*{Funding Sources}
No other outside source of funding was provided.
\pagebreak{} 
%
%
%
%
%


\chapter*{NOMENCLATURE}
\addcontentsline{toc}{chapter}{NOMENCLATURE}  



\hspace*{-1.25in}
\vspace{12pt}
\begin{spacing}{1.0}
	\begin{longtable}[htbp]{@{}p{0.35\textwidth} p{0.62\textwidth}@{}}
		CPU	&	Central Processing Unit	\\ [2ex]
		GPU		&	Graphics Processing Unit \\	[2ex] 
		TPU			&	Tensor Processing Unit \\	[2ex]
		ML & Machine Learning \\ [2ex]
		LLM & Large Language Model \\ [2ex]
		IP & Instruction Pointer \\ [2ex]
		L1D & Level 1 Data Cache \\ [2ex]
		L1I & Level 1 Instruction Cache \\ [2ex]
		L2C & Level 2 Cache \\ [2ex]
		LLC & Last Level Cache \\ [2ex]
		KV Cache & Key Value Cache \\ [2ex]
		RAM & Random Access Memory \\ [2ex]
		IoT & Internet of Things \\ [2ex]
		LRU & Least Recently Used \\ [2ex]
		NRU & Not Recently Used \\ [2ex]
		RRIP & Re-reference Interval Prediction\\ [2ex]
	\end{longtable}
\end{spacing}

\pagebreak{}

%
%
%
%
%

\phantomsection
\addcontentsline{toc}{chapter}{TABLE OF CONTENTS}  

\begin{singlespace}
\renewcommand\contentsname{\normalfont} {\centerline{TABLE OF CONTENTS}}

\setcounter{tocdepth}{4} 

\setlength{\cftaftertoctitleskip}{1em}
\renewcommand{\cftaftertoctitle}{%
\hfill{\normalfont {Page}\par}}

\tableofcontents

\end{singlespace}

\pagebreak{}


\phantomsection
\addcontentsline{toc}{chapter}{LIST OF FIGURES}  

\renewcommand{\cftloftitlefont}{\center\normalfont\MakeUppercase}

\setlength{\cftbeforeloftitleskip}{-12pt} 
\renewcommand{\cftafterloftitleskip}{12pt}

\renewcommand{\cftafterloftitle}{%
\\[4em]\mbox{}\hspace{2pt}FIGURE\hfill{\normalfont Page}\vskip\baselineskip}

\begingroup

\begin{center}
\begin{singlespace}
\setlength{\cftbeforechapskip}{0.4cm}
\setlength{\cftbeforesecskip}{0.30cm}
\setlength{\cftbeforesubsecskip}{0.30cm}
\setlength{\cftbeforefigskip}{0.4cm}
\setlength{\cftbeforetabskip}{0.4cm}



\listoffigures

\end{singlespace}
\end{center}

\pagebreak{}

%
\phantomsection
\addcontentsline{toc}{chapter}{LIST OF TABLES}  

\renewcommand{\cftlottitlefont}{\center\normalfont\MakeUppercase}

\setlength{\cftbeforelottitleskip}{-12pt} 

\renewcommand{\cftafterlottitleskip}{1pt}

\renewcommand{\cftafterlottitle}{%
\\[4em]\mbox{}\hspace{2pt}TABLE\hfill{\normalfont Page}\vskip\baselineskip}

\begin{center}
\begin{singlespace}

\setlength{\cftbeforechapskip}{0.4cm}
\setlength{\cftbeforesecskip}{0.30cm}
\setlength{\cftbeforesubsecskip}{0.30cm}
\setlength{\cftbeforefigskip}{0.4cm}
\setlength{\cftbeforetabskip}{0.4cm}

\listoftables 

\end{singlespace}
\end{center}
\endgroup
\pagebreak{}  


%
%


\pagestyle{plain} 
\pagenumbering{arabic} 
\setcounter{page}{1}


\chapter{\uppercase {introduction and literature review}}
\section{Background}
Machine Learning (ML) algorithms, particularly Large Language Models (LLMs), have become a valuable tool in a variety of disciplines. Their robust architecture makes it easy to train for a specific and complex task that would normally require a human or a specialized algorithm to do. The resilience of these models has led them to be used even in cache design \cite{cache-deep-learning}, and by scaling up model size, LLMs are able to quickly adapt to new inputs with little to now training on the specific task \cite{zeroshot}.

At a high level, an ML algorithm utilizes previous training and some input to produce an output. As an example, a model could be trained with a set of labeled images, given unlabeled images, and asked to infer their labels. The important takeaway is that the accuracy of the output is determined more so by the quality of the training data and the models ability to use said data, rather than the exact architecture used by the model.

LLMs have two broad types: encoders and decoders. Encoders a set of tokens and produce an encoding, which could be used for analysis or an input for another model. An example of an encoder is BERT \cite{bert}, which we had originally used to test the LLM's CPU performance by using it to perform sentiment analysis, determining whether a given token has a positive or negative sentiment. Decoders produce an output from their tokens, most commonly text. A well known decoder is Open AI's GPT4 model \cite{openai2024gpt4technicalreport}, and for this paper we have used QWEN \cite{qwen}.

Decoders can be broken into two phases, a Prefill phase and a Decode phase which are compute bound and memory bound respectively \cite{llm-cpu-perf}. The Prefill phase will take a given input and produce an initial token. The Decode phase iteratively generates new tokens, using previously generated tokens as its input \cite{llm-cpu-perf}. When decoding, a token depends on every previous token \cite{nomad}, which must be calculated each iteration unless there is a KV Cache. 

The KV Cache is an optimization aimed at reducing inference times by caching the Key Value pairs of token embeddings, reducing the time complexity of decoding from $O(n^2)$ to $O(n)$, but introducing memory overhead \cite{llm-cpu-perf}. The KV Cache has become essential for larger models, and it was noticeable when running my own experiments, but the memory overhead exacerbates the issues of the memory bound decoder phase: more data must be accessed from memory. It is important to note that the KV Cache is not a physical piece of hardware, but instead a space to cache these pairs.

\subsection{Advantages of an LLM on a CPU}
The desire to run an LLM on a CPU ultimately stems from some limitation that prevents the model from running with an accelerator such as a GPU or TPU. An accelerator is an additional constraint placed on a system that wishes to use LLM to run inference, which may not be possible or preferable. Conflicting constraints could be related to budget, energy consumption, privacy, or security. By increasing the performance of an LLM in a CPU environment, the accelerator constraint will lessen and the accessibility of LLMs will increase.

With the introduction of a KV cache to speed up inference and increased model sizes, memory requirements are now a concern for accelerators \cite{llm-cpu-perf}. GPUs have a relatively low memory capacity compared to what a CPU can have access to. Work has been done to offload parts of the KV cache to RAM but the latency overhead is undesirable \cite{llm-cpu-perf} \cite{cpu-gpu-orch}. Memory is quickly becoming the main bottleneck to LLM performance\cite{memorywall}, which CPUs will be able to better handle with efficient memory management via their cache system as long as improvements can be made to their ability to execute these models.

An advantage that may not be immediately considered is data security. CPUs have the ability for security in their hardware that GPUs lack, with an example being SGX. SGX acts on the hardware level to create a container that guarantees the "confidentiality and integrity" of the data within \cite{sgx}. This could be an important consideration for those utilizing data centers for their models, as this blocks unauthorized software that could be running on a machine from interacting with the model in a malicious way, such as taking or modifying the model's data. 

Most processes run by data centers have little to no need for accelerators, and the environments in a data center are tailored towards CPU usage. As LLMs become more popular, their usage in cloud services and data centers have increased, as well as the number of accelerators to handle these \cite{llm-cpu-perf}. If smaller models could run inference without an accelerator and CPUs could supplement the processing power of GPUs by handling some of the model layers \cite{llm-cpu-perf} \cite{cpu-gpu-orch}, these services could achieve better resource utilization.

Energy cost is a concern when a system has significant power constraints, such as relying on battery power or energy harvesting. This would be most prominent in edge computing and IoT, with TinyML focusing on the use of small machine learning models to consume milliwatts of power \cite{tinyml-review}. Currently, CPUs have less energy efficiency than a GPU, but have a lower average power consumption \cite{power-consumption}. A GPU will draw more power than a CPU, but due to the inefficient execution of the model on a CPU, the run time is much larger, and by extension the total energy cost. This is much less an advantage and more a trade-off, but for an environment with reliable power flow, using a CPU instead of a GPU could cut the average power consumption by half \cite{power-consumption}.

For systems with no accelerator, their options for running LLM inference are limited. One option is to leverage resources in the cloud to perform this task, but this exposes the data to external servers, raising privacy concerns. Locally hosting an LLM offers increased privacy and the ability to adjust the model with confidential data \cite{cpu-gpu-orch}. Privacy is especially important for Internet of Things (IoT) devices that rely on machine learning for tasks like speech recognition \cite{tiny-ml}, as transmitting sensitive user data over the internet increases the risk of interception or misuse. These resource constrained environments would benefit from being able to use a CPU to enable or supplement the ability to run an LLM locally.

\subsection{Why an LLM on CPU is Uncommon}
With these advantages, why is running inference on an LLM with a CPU uncommon? The memory wall faced by LLMs has only recently become a major issue as the model sizes exceed GPU memory, and CPUs execute models inefficiently. Improvements must be made to their execution in order to make use of the advantage of the CPU's features and overcome the memory wall.

One of the largest hurdles to overcome when running LLMs on a CPU is the time requirement. LLMs utilize many matrix operations that benefit greatly from parallelization. The performance of a model on a CPU is highly dependent on multithreading \cite{power-consumption} and multiple cores \cite{cpu-vs-gpu-dl}. Without an accelerator, which specializes in this task, the inference time is substantially increased as every matrix operation is performed sequentially instead of in parallel, with the GPU being 4-5x faster \cite{cpu-vs-gpu-dl}. For smaller models, such as those that would be used in TinyML and those with small batch sizes, the performance is comperable since the number of parallelizable operations has decreased \cite{cpu-vs-gpu-dl}, but this gap remains a large issue. Efforts are being made to reduce \cite{nomad} or accelerate \cite{llm-cpu-perf} these operations.

\subsection{Why Focus on Caches?}
With the need to improve the CPUs performance, we recall the two phases of a decoder model: Prefilling and Decoding. The memory-bound Decoding phase increases with the number of tokens that the model will create \cite{nomad}. The exact tranlsation from tokens to words varies, but when using 128 tokens in my own experiments we recieved responses ranging from 75-110 words. More complex models may use thousands of tokens or more. With this in mind, the decoding phase is the main bottleneck of performance for a CPU.

The primary concern when improving a memory-bound process is memory efficiency. Reducing the delay that it takes for an address to be retrieved from memory is key. In the scope of computer architecture, this could be done by making an adjustment to the caches of a CPU. Adjusting the memory capacity of a cache is expensive and unlikely to scale with the demands of models. Instead, using the space within a cache as efficiently as possible is the key. This is done through two ways: prefetching and replacement policies, which are discussed further in the related work.

\section{Related Work}
The concept of running LLM inference on a CPU has only recently started to be explored and there are many avenues of improvement. Closely related to this concept is the field of TinyML, which has similar goals of using ML algorithms in efficient ways. TinyML differs in that it is primarily focused in IoT and small edge computing, though making improvements to LLM performance on a CPU could benefit this area. Finally, to better understand the behavior shown in the experiments, we discuss the prefetchers and cache replacement policies used. 

\subsection{LLM CPU}



LLMs traditionally use accelerators such as GPUs and TPUs for inference, but larger models that exceed the accelerator's memory capabilities due to their model parameters and KV cache size pose a challenge to run. An attempted solution to this issue is memory offloading, but this comes with the overhead associated with data transfer between the CPU and accelerator \cite{llm-cpu-perf}\cite{cpu-gpu-orch}. Quantization, a technique often used on weights to reduce their memory footprint, could be used to offset this issue, but reduces the quality of the model \cite{cpu-gpu-orch}. CPUs have the memory capacity to run these models due to the higher memory capacity, and can outperform a GPU in these circumstances \cite{llm-cpu-perf}. 

Because the decoding process is memory bound, the focus on performance in this area is efficient memory management \cite{llm-cpu-perf}. To improve the decoding speed, caches must have a high hit rate, which can be done with a prefetcher or replacement policy that accommodates the memory pattern of the model. 

Another area of investigation is the execution of a model using the CPU and GPU together in an attempt to reduce the overhead caused by the resource transfer performed when memory offloading \cite{llm-cpu-perf} \cite{cpu-gpu-orch}. This method is most effective with a small batch size, where the CPU can run layers during inference \cite{cpu-gpu-orch}. Even in situations where a CPU will not outperform a GPU, utilizing CPU resources to run inference can alleviate the demand for GPU resources in data centers \cite{llm-cpu-perf}. Environments such as these would benefit from increased performance when running an LLM on a CPU, even if accelerators are available. Since these layers would be involved in the decoding process, they would be memory bound and would benefit from more efficient memory usage.

Most of the time spent running inference is with multiply-add operations, which are computed between every pair of tokens. By creating a lookup table for the dot-products and the multiply-add operation can be avoided entirely. However, this creates an overhead from looking up the memory, which is solved by compressing the look-up table into a SIMD register \cite{nomad}.


\subsection{TinyML}

TinyML is a discipline of machine learning with the focus on using very small models for environments with high hardware constraints, either with power and resources. TinyML is designed to be used for hardware that draws 1 mWatt or less, with use cases such as speech or gesture recognition, wake words, or image recognition \cite{tiny-ml-benchmark} \cite{tiny-ml}. Currently, the hardware used for TinyML varies significantly, and a suitable benchmark is needed to standardize its efficiency \cite{tiny-ml-benchmark}.

The goals of this research and TinyML are similar despite covering different scopes. By making improvements to an LLM's performance on the CPU, we open the door for more complex models to be used in an environment where it was previously infeasible to. An IoT device could determine if a more complex model is needed to process the information, \cite{tinyml-review} and rather than transmitting the data to the cloud, the data could be sent to a locally running model, preserving privacy and reducing transmission times. Additionally, because models are memory bound \cite{nomad} \cite{llm-cpu-perf} \cite{memorywall}, cache architecture built with machine learning in mind will reduce the amount of time the CPU spends waiting for memory, increasing energy efficiency.

\subsection{Cache Prefetchers}

Prefetching is a concept in which caches attempt to predict the next required address using some sort of algorithm or heuristic and request it prior to the CPU needing the address, reducing the amount of stalling and speeding up the process of accessing memory. The most simple prefetching algorithm grabs the next line of addresses using the current address as a point of reference. There are three different types of prefetchers: stride prefetching, spatial prefetching, and temporal prefetching. 
Temporal prefetching attempts to prefetch data by identifying temporal correlations in cache misses and replaying these misses when they are found again \cite{temporal}. However, these prefetchers require large amounts of metadata to function and introduce an overhead \cite{temporal}. In the context of an LLMs execution, a temporal prefetcher is unlikely to provide a significant advantage to stride or spatial prefetching since an LLM lacks high temporal locality.

Stride prefetching attempts to calculate the stride, or delta, of addresses and use this to predict what the next address will be. A simple implementation of this uses a stride prediction table to calculate the stride of memory accesses in arrays during looping \cite{stride}. Prefetchers in this category may find success in the LLMs repeated weight and KV cache accesses with each token, but the prefetcher must be sophisticated enough to determine these addresses and have enough confidence to start prefetching those addresses. If we notice a decrease in performance, the prefetcher is predicting the wrong deltas, as a misprediction is more expensive than making no guess.
Berti, the winning prefetcher of the 3rd Data Prefetching Championship makes use of the idea the best timeliness will be found with the best delta, and that this varies from page to page. Berti has two sets of data that it captures: data used to make predictions and data used to evaluate confidence with a prediction \cite{berti}. Because confidence is determined by a series of matches dependent on the instruction pointer and accessed pages \cite{berti}, Berti can avoid irregular memory patterns which prevents mispredictions. For pages that have not been accessed in some time, Berti uses a burst mode which prefetches every memory block from the first access in that page to the calculated stride \cite {berti}. 

The Instruction Pointer Classifier based Prefetcher (IPCP) uses instruction pointers to determine the stride of a memory access pattern. IPCP divides an instruction pointer stride into three different classifications: constant, complex, and global stream. Constant strides are characterized by a repeated stride, with a 2 bit counter for cases where a stride might be slightly offset from the rest. Complex strides record a signature of accesses to handle a situation where a simple stride counter would mispredict or never gain enough confidence to make predictions. Finally, IPCP has a global stream for a series of memory accesses that are accessed independently of the instruction pointers, using an initial IP as the trigger \cite{ipcp}.

The performance of Berti and IPCP is contingent on a process having a predictable pattern of memory accesses. The performance of a model will depend on the memory access patterns of the model and the framework that implements that model. Stride prefetchers will perform best when weights and other data required for inference are stored and accessed contiguously with a consistent stride. Since an LLM will perform with arrays of data, these could be very promising for improving memory efficiency.

Bingo is a spatial prefetcher that uses multiple spatial heuristics categorized into different tables to store an event, a length, and a prediction associated with an event. Events can be split into two categories: long and short, with long events having a high accuracy but low occurrence, and short events having a lower accuracy but a high occurrence \cite{bingo}. This could be promising for an LLM if layer accesses can be categorized into a long event, since each layer would be accessed once per token.

With a large set of memory in use, LLMs will use many pages. Each page is only 4 kB, with some LLMs surpassing 40 GB. A prefetcher able to separate its prediction page by page, such as Berti, is an improvement from prefetchers that have only a global stride, but if a prefetcher could work through page boundaries, there could be additional improvements

The Signature Path Prefetcher (SPP) makes use of a short history, tracking strides, to create a prediction. The prediction, along with the history, is put together to create a signature which is used to make another prediction, which is used to repeat this process, This creates a path of predictions whose length varies depending on a measure of confidence in the predictions. Page boundaries would present an issue for SPP since the history is lost when moving pages. However, by using a global history register, SPP is able to store the signature that crosses the page boundary and continue making predictions with no warmup time. \cite{SPP}

\subsection{Cache Replacement Policies}

Caches have a limited space and must make evictions when full. This introduces the problem of choosing which data to evict, since an arbitrary eviction could remove data that needs to be accessed shortly thereafter. The most simple eviction method is Least Recently Used (LRU), which assumes that memory access patterns will have a high temporal locality. It follows that whatever address was accessed the least recently will be needed the least compared to every other address, so it is evicted.

For an LLM this is inefficient, as all of the memory accesses in an LLM do not necessarily have a high temporal locality. Additionally, the working set of an LLM will easily exceed the cache capacity, further decreasing the LRU performance. The current token in a decoding process depends on every prior token, resulting in many references to older data. Older tokens, despite the fact that they are frequently reused, may be evicted in favor of keeping data that will be used once or infrequently accessed. Finding a cache replacement policy capable of exploiting this behavior could improve performance. 

SRRIP and DRRIP are designed with the above principle in mind: some applications have distant references, generally with working sets larger than the cache and with "frequent bursts of references to non-temporal data" \cite{drrip}. This described behavior is very similar to what occurs in memory as an LLM executes. These two cache prefetchers rely on a Re-reference interval prediction (RRIP), which is classified into near-immediate and distant based on a Not Recently Used (NRU) policy.

NRU is an approximation of LRU that assigns one bit to each block in the cache. When a cache miss occurs, NRU will evict blocks that have a distant re-reference value. If it cannot find a block, the policy updates every block to have a distant future and performs the eviction \cite{drrip}. Blocks that are not re-referenced will remain distant, and those that get re-referenced are set back to having a near-immediate value. However, by always predicting that a block will have a near-immediate re-reference interval, blocks that actually have a distant re-reference interval will not be evicted until the update occurs, and the policy may incorrectly evict another address first. Since every block is set to having a distant re-reference interval, a tie breaker evicts the leftmost block, which may not be the optimal choice.

This issue is what RRIP aims to solve by predicting the re-reference interval by using more information than just one bit. On a cache hit, RRIP sets the re-reference interval to near-immediate. On a miss, the re-reference intervals are increased incrementally until a distant block is found and evicted. This creates wiggle room for blocks that, while they may be least recently accessed, have a higher probability of being re-referenced than a block that just entered the cache. The Static Re-reference Interval Prediction (SRRIP) is RRIP using a static policy determined by cache hit and misses. Dynamic Re-reference Interval Prediction (DRRIP) improves a shortcoming of SRRIP by dynamically deciding when to use SRRIP and a thrash-resistant Bimodal RRIP to prevent issues \cite{drrip}. An difference in performance between DRRIP and SRRIP would show that the re-reference interval for all blocks exceeds the capacity of the cache.

The Signature-based Hit Predictor (SHiP) predicts re-reference intervals by associating a signature with a re-reference prediction, seeking to make improvements from RRIP. SHiP makes use of memory regions, the program counter, and instruction sequences to learn a signature and make predictions based on the re-reference interval of that signature. Since the re-reference prediction is signature based, when an insertion matches a signature, the associated re-reference interval can be used. Importantly, SHiP still uses SRRIP as its actual replacement policy, but determining which lines are distant or near-immediate is done with the signature system \cite{SHip}.

By aiming to predict the re-reference interval of a cache line as it enters, SHiP should be able to accommodate for the characteristics of an LLM. SHiP should be able to identify which data is used every layer, and which data is used once a layer. As each token in the decoding phase is generated, the same signatures should be used and SHiP would be able to capitalize on this by already knowing the re-reference patterns.

Ultimately, the best-performing cache replacement policy is one that can accurately capture a single layer of the model and utilize that behavior for subsequent layers while tolerating differences that may arise as prior tokens are used for input. The re-reference intervals of these may be large, and for intervals large enough to warrant it, it may be a wise idea to implement cache bypassing since they are utilized only once \cite{cache-bypass}.
%
%



\chapter{\uppercase {methodology}}

\section{Experimental Setup and Replication}

This section covers in detail how we achieved the results found in the next chapter. The first section, Tracing Methodology, covers the groundwork needed to run the experiments detailed in the following section. At a high level, Intel Pin and ChampSim's tracer tool were used to collect a memory profile of a binary, which was then simulated in ChampSim with different cache implementations. The access patterns reflected in a default cache architecture were recorded and used for the memory analysis.

This work produced the information necessary to run the two experiments: seeing what current prefetchers and replacement policies would work best for a given LLM model and what useful patterns the memory footprint could reveal based on access frequency and total access counts. These two should provide useful information on how to improve the cache architecture for LLM inference speed by showing what improves performance and why.

\subsection{Tracing Methodology}

ChampSim \cite{champsim} requires a memory trace to perform a simulation. This is done using an Intel Pin tool provided by ChampSim to collect information about a binary. The instruction pointer and information about branching, registers, and memory operands are collected. Each instruction traced takes 64 bytes, which adds up quickly for large traces, so the trace is compressed using xz, which can be read by ChampSim. The compression is significant, with the roughly 453 gigabyte file being compressed to 200 megabytes.

Understanding Intel Pin is important for making adjustments to the tracing tool, which must be done to perform the experiments. Intel Pin is a just-in-time compiler that uses programs called tools to get information from a binary. A main function is used to declare a set of callback functions that trigger on certain events in the binary, such as a thread starting, an executable loading, or when a new instruction is reached. The main feature used by the trace is the ability to insert function calls before an instruction is executed, allowing information about the instruction to be read and analyzed. Once these callback functions are declared, the binary execution begins.

The ChampSim tracer uses Intel Pin to run code on every instruction, starting the process by using a global variable to collect the current instruction pointer. If the instruction is a branch, appropriate branch information is collected before the register reads, register writes, memory reads, and memory writes are collected. This provides a complete picture of the memory footprint in a binary.

Because the purpose of this trace is to analyze the performance of an area of interest in a binary instead of analyzing the performance of a cache implementation, the options available with the default tracer are insufficient for the needs of the experiment. Being able to skip instructions and specify an amount of instructions to trace is not enough control, since one would have to count the number of instructions before the area of interest. Additionally, we noticed some undefined behavior with the tracer that raised some questions.

The first issue was solved by inserting a function before the tracer's normal logic that checks if we are in an area of interest and, if so, allows for the rest of the tracer code to execute. This same function will check if the area of interest has been exited to disable the instrumentation, ending the trace. The next issue took some digging to understand, but ultimately had to do with multi-threading. ChampSim's trace formant and, by extension, the tracer are unable to handle multithreading. At first, we believed this to only effect the stopping condition for the tracer, so we implemented synchronization into global variable access and made the information collection thread specific. However, even a safe tracer would mangle the results since ChampSim can only handle one thread, so the binary must only run a single thread. An improved version of the tracer should throw an error or issue a warning if a new thread is created to prevent the issue that we ran into.

\subsection{Simulation Methodology}

ChampSim is a trace-based CPU simulator focused on cache architecture \cite{champsim}. Because ChampSim relies on a trace instead of emulating a platform and running the binary directly, the scope of what ChampSim handles is smaller than other simulators, but the execution speed is noticeably faster. Recent versions of ChampSim function by generating a cache structure specified by a .json configuration, then run a provided trace. This process can be seen in Figure \ref{fig:champ_commands} Two numbers of instructions are specified: instructions that are simulated but not tracker, and simulated instructions that record their statistics. The purpose of skipping instructions is to allow the caches to warm up and to avoid recording compulsory misses. Once the trace has been completed, the cache statistics are printed. For the concerns of our memory-bound program, a combination of the miss rate, IPC, and prefetch traffic is what determines the performance of a policy.

ChampSim offers the ability to implement custom prefetching and replacement policy logic and to adjust properties of the caches. Figure \ref{tab:cache_param} shows the configuration of the caches used, which are constant across the different simulations for the prefetchers and replacement policies.

\begin{figure}[h]
\begin{small}
\begin{verbatim}
ChampSim:
$ ./config.sh ./champsim_config.json
$ bin/champsim --warmup_instructions 200000000 
  --simulation_instructions 7044850000
  ../llm_trace.trace.xz >../out_stats.txt
\end{verbatim}
\end{small}
\caption{Examples of ChampSim Commands}
\label{fig:champ_commands}
\end{figure}

\begin{table}[h]
    \centering
    \begin{tabular}{|c|c|c|c|c|c|c|}
         \hline
         Cache & Cache Size & Sets & Ways & RQ Size & WQ Size & MSHR Size \\
         \hline
         L1I & 32kB & 64 & 8 & 64 & 64 & 8 \\
         \hline
         L1D & 48kB & 64 & 12 & 64 & 64 & 16 \\
         \hline
         L2C & 512kB & 1024 & 8 & 32 & 32 & 32 \\
         \hline
         LLC & 4096kB & 4096 & 16 & 32 & 32 & 64 \\
         \hline
    \end{tabular}
    \caption{Cache Parameters}
    \label{tab:cache_param}
\end{table}

\subsection{Memory Analysis Methodology}

ChampSim's simulations can show the performance of a given cache implementation, but additional analysis is needed to understand the patterns in memory and determine what can be exploited by a current or new prefetcher and replacement policy. The tensors of the model must be mapped to the addresses used, and those addresses should be checked for a pattern. Addresses on the extreme ends, those that are frequently used or used only once, should be investigated as well.

To perform this analysis, preparations need to be made prior to compilation which will ensure the binary has the necessary information for debugging and that the data is consistent. The binary must be compiled with all libraries statically included and it must be compiled in debug mode to include symbol information. This provides all of the information necessary to start the analysis, but one more step is needed.

The memory footprint of a program is represented in a virtual address space that is determined at link time. One can use objdump to see an example of this, showing function and constant addresses during this. However, to increase the security of a system this virtual address space is randomized on each execution. This is done with two methods: Address Space Layout Randomization (ASLR) and Position Independent Execution (PIE). These are controlled at two different levels of the system and interact with the program at different points.

PIE is implemented during compilation and must be changed using a flag when building the program. PIE will adjust all program addresses to be in a relative position to one another instead of an absolute position in the virtual address space. When disabled, the global variables, constants, and other program data will be constant across executions. However, these adjustments do not impact the stack or the heap. To keep the stack and heap relatively consistent through executions, ALSR must be disabled prior to the execution of the program, and is changed for a process or globally for the operating system. With both of these features disabled, the next challenge arises.

Even with a constant stack and heap location, the stack and heap can still vary between executions. Before the first variable is loaded into the stack, there can be a number of arguments or environmental variables passed to the program. This adjusts the address of every variable on the stack, but their offsets remain the same. The heap is similarly affected, in that the allocator determines where an address on the heap will go. This behavior could be consistent, but is not guaranteed.

There are a few ways to overcome this obstacle. The first is to utilize a global variable or other constant address to determine where the stack starts. If an address is assigned to a global variable pointer, the instruction corresponding to that operation can be found, and other variables may be found using that address and the relative offsets or source code. The next option is to use the instruction address directly to determine where a variable is located. The binary must be disassembled to find where the instruction address is in the source code, then the address can be found. Since the area of interest lies outside of any global variable declarations, the second option must be used.

We used ChampSim to generate the data used for the memory analysis since this would provide an accurate representation of how the memory was flowing through the caches. ChampSim's caches are connected to each other and the CPU through data channels, which channels transmit requests for data packets. By writing the data flowing through read requests for the the CPU's L1I and L1D channels to file, a complete footprint of the memory can be obtained. To capture the LLC's requests, the try\_hit function was modified. These files printed the address and the cycle in which the address was accessed.


We used GDB to disassemble the binary and help in the analysis. GDB can be used to examine the source code side by side with a disassembly of the binary, allowing one to easily correlate an IP to its source code. Additionally, it can step through the execution of the binary and determine the memory layout of a given variable, allowing us to view how the tensors and model layers are placed in the stack and heap. This can be used to determine which addresses in the L1D output belong to which variables, as stack and heap addresses appear in two different memory bands. Heap-allocated addresses, while inconsistent across multiple executions, tend towards being allocated in the same area, so their most significant bits are the same.

\section{Overview of Experiments}

To carry out the experiments, we compile Llama.cpp with the required settings for the memory analysis. Llama.cpp is a framework used for conducting inference with different back-ends and allows for inference to occur with only a CPU. The specific binary used was the llama-cli example, allowing for interaction with a model through the command line. Modifications to the binary were made to force interaction mode off to allow a trace to occur without user input. We used the QWEN decoder model with 0.5B parameters and generated 128 tokens using one thread, Figure \ref{tab:model_params} provides further details. Figure \ref{fig:llama_command} shows how these parameters are used. The trace starts after the model has been loaded and has gone through a warm-up run and ends once the decoder has fully generated its response. This traced around seven billion instructions that were compressed to be used in ChampSim.

\begin{figure}[h]
\begin{small}
\begin{verbatim}
$ ./llama-cli-nopie-new -m /qwen2.5-0.5b-instruct-fp16.gguf 
  --prompt "I believe the meaning of life is" -n 128 -t 1
\end{verbatim}
\end{small}
\caption{Llama.cpp Command}
\label{fig:llama_command}
\end{figure}

\begin{table}[h]
    \centering
    \begin{tabular}{|c|c|c|c|}
         \hline
         Parameter & Value  \\
         \hline
         Model Parameters & 0.5 Billion \\
         \hline
         Threads & 1 \\
         \hline
         Input & "I believe the meaning of life is" \\
         \hline
         Tokens & 128 \\
         \hline
         Context Size & 4096 \\
         \hline
         Batch Size & 2048 \\
         \hline
         Floating Point Size & 16 bits \\
         \hline
    \end{tabular}
    \caption{Model Parameters}
    \label{tab:model_params}
\end{table}

\subsection{ChampSim Simulations}

By examining how current prefetchers and replacement policies perform with an LLM model, insights could be gained based on their relative performance to one another, exposing a pattern that could be exploited in a new implementation. To achieve this, the QWEN trace is simulated with various configurations, each adjusting the prefetchers or replacement policies of the caches. The trace captured the prefill and decoding phases, but previous work \cite{llm-cpu-perf} shows that most memory operations will occur in the decoding phase. Since the uncompressed format is 64 bytes per instruction, we wrote a program to split the binary where the decoding phase starts. We noticed that the first output occurred at around 5.8 billion instructions, with the IPC dropping at around 5.7 billion instructions. 200,000,000 instructions were skipped as a warm-up, resulting in 1,344,850,000 simulated instructions recorded in the decoding phase. Table \ref{tab:sim_instr} shows how the phases are split. Simulations will focus on the decoder phase, as that is where efficient memory management will have an impact. Table \ref{tab:sim_vars} shows which prefetchers and replacement policies will be used and which versions of the simulator they run on.

\begin{table}[h]
    \centering
    \begin{tabular}{|c|c|}
         \hline
         Phase & Instructions Captured  \\
         \hline
         Total & 7,244,850,000 \\
         \hline
         Prefill & 5,700,000,000 \\
         \hline
         Decoder Phase & 1,544,850,000 \\
         \hline
    \end{tabular}
    \caption{Overview of Simulation Parameters}
    \label{tab:sim_instr}
\end{table}

\begin{table}[h]
    \centering
    \begin{tabular}{|c|c|}
         \hline
         Prefetcher & Replacement Policy \\
         \hline
         Next Line & LRU \\
         \hline
         IP Stride & SRRIP \\
         \hline
         Berti & DRRIP \\
         \hline
         Bingo & SHiP \\
         \hline
         Bouquet (IPCP) & \\
         \hline
         SPP & \\
         \hline
    \end{tabular}
    \caption{Prefetchers Used in Simulation Experiment}
    \label{tab:sim_vars}
\end{table}




\subsection{Trace Analysis}

While comparing simulations provides some insight, analyzing the memory trace and observing the memory footprint directly will provide additional context to the performance of those simulations and potentially reveal useful patterns. There are two data points to be gathered in the trace analysis: a histogram of memory accesses and a count of accesses for every address. By printing out the memory address and cycle for the L1D cache, both can be gathered.

The access counts are crucial for analyzing the memory footprint, since they show which addresses are outliers. Addresses with high access counts or with a notable number, such as the number of tokens generated, are prime targets to be investigated first. These can be found in the dump of L1D accesses along with the cpu cycle showing when the data was requested. By comparing this cycle to the dump of L1I accesses, we can approximate the instruction that called the data and use GDB to determine which variable in the source code the address corresponds to.

This alone is not enough to make a confident guess. Additional context, such as data accessed around our address of interest, whether the address was heap or stack allocated, address offsets, the data structure in the source code, and the most significant bits of an address, is needed to increase the confidence that an address corresponds to a variable.

By finding a meaningful address, such as a weight or logit, we can find the delta in token executions and print a histogram of this region. Because the decoding process will have little deviation from token to token, a graph of this delta should be mostly accurate for the entire decoding process. The magnitude of this delta should also provide information on how wide the working set is for a cache replacement policy and prefetchers.

%
%



\chapter{\uppercase {results}}

\section{Simulation Results}

Tables \ref{tab:pref_l1d}, \ref{tab:pref_l2c}, \ref{tab:pref_llc} show the performance of the prefetchers used in various caches. The L1I was not investigated because it is not related to our topic of interest. Bingo has the lowest miss percentage and the highest ratio of useful prefetches. The L2C cache had the highest miss rate, though next line prefetching can cut the miss rate by half and had the best performance. The ratio of useful prefetches is also very high. IPCP performed abnormally poor relative to the other prefetches. The LLC cache had the lowest miss percentage of the three caches. The next line prefetcher degraded performance, implying that the stride of an LLC could be complex, or that there are many jumps between pages.

Table \ref{tab:repl} shows the replacement policies used. The LLC is often not prefeched, so these provide a more accurate look at what could be done to improve LLC performance. With 4 MB, the working set appears to be able to be contained in the cache while decoding occurs, resulting in a low miss percentage of 0.065\%. DRRIP was able to reduce the miss percentage further to 0.018\%.


\begin{table}[h!]
    \centering
    \begin{tabular}{|c|c|c|c|c|c|}
         \hline
         Prefetcher & IPC & Miss Percentage & Total Accesses & Useful Pref. & Pref. Issued \\
         \hline
         None & 1.47 & 4.056\% & 477321401 & 0 & 0 \\
         \hline
         Next Line & 1.616 & 3.204\% & 477327484 & 4222503 & 140286827 \\
         \hline
         Berti & 1.615 & 3.266\% & 477316442 & 4219198 & 285665974 \\
         \hline
         Bingo & 1.622 & 3.176\% & 477322558 & 4171697 & 4282771 \\
         \hline
         IPCP & 1.616 & 3.010\% & 477314472 & 4225845 & 194604946 \\
         \hline
    \end{tabular}
    \caption{Prefetchers Used on the L1D Cache}
    \label{tab:pref_l1d}
\end{table}

\begin{table}[h!]
    \centering
    \begin{tabular}{|c|c|c|c|c|c|}
         \hline
         Prefetcher & IPC & Miss Percentage & Total Accesses & Useful Pref. & Pref. Issued \\
         \hline
         None & 1.47 & 69.884\% & 10807473 & 0 & 0 \\
         \hline
         Next Line & 1.58 & 30.786\% & 10807472 & 4227792 & 4343668 \\
         \hline
         IP Stride & 1.584 & 32.122\% & 10807473 & 4081058 & 12035800 \\
         \hline
         Berti & 1.578 & 30.848\% & 10807474 & 4218848 & 62242118 \\
         \hline
         Bingo & 1.586 & 31.247\% & 10807472 & 4178147 & 4194772 \\
         \hline
         IPCP & 1.47 & 69.853\% & 10807474 & 3790 & 14877 \\
         \hline
         SPP & 1.555 & 41.087\% & 10807472 & 3112729 & 61874552 \\
         \hline
    \end{tabular}
    \caption{Prefetchers Used on the L2C Cache}
    \label{tab:pref_l2c}
\end{table}

\begin{table}[h!]
    \centering
    \begin{tabular}{|c|c|c|c|c|c|}
         \hline
         Prefetcher & IPC & Miss Percentage & Total Accesses & Useful Pref. & Pref. Issued \\
         \hline
         None & 1.47 & 0.065\% & 10789370 & 0 & 0 \\
         \hline
         Next Line & 1.464 & 0.399\% & 10789371 & 4891 & 4342383 \\
         \hline
    \end{tabular}
    \caption{Prefetchers Used on the LLC Cache}
    \label{tab:pref_llc}
\end{table}

\begin{table}[h!]
    \centering
    \begin{tabular}{|c|c|c|c|}
         \hline
         Repl. Policy & IPC & Miss Percentage & Total Accesses \\
         \hline
         LRU & 1.47 & 0.065\% & 10789370  \\
         \hline
         SRRIP & 1.47 & 0.021\% & 10789370 \\
         \hline
         DRRIP & 1.47 & 0.018\%  & 10789371 \\
         \hline
         SHiP &  1.47 & 0.037\% & 10789371 \\
         \hline
    \end{tabular}
    \caption{Cache Replacement Policies for the LLC Cache}
    \label{tab:repl}
\end{table}

\pagebreak

\section{Memory Analysis Results}

\begin{figure}
\centering
\includegraphics[width=\textwidth, trim={0 0 0 0.72cm}, clip]{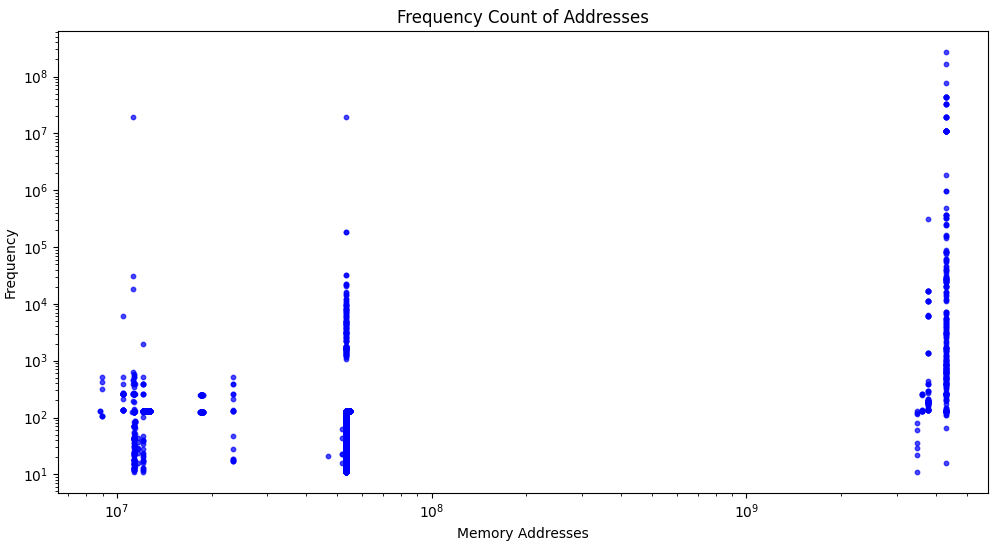}
\caption{Address vs Number of Times Accessed - Full}
\label{fig:freq_plot_full}
\end{figure}

\begin{figure}
\centering
\includegraphics[width=\textwidth, trim={0 0 0 0.70cm}, clip]{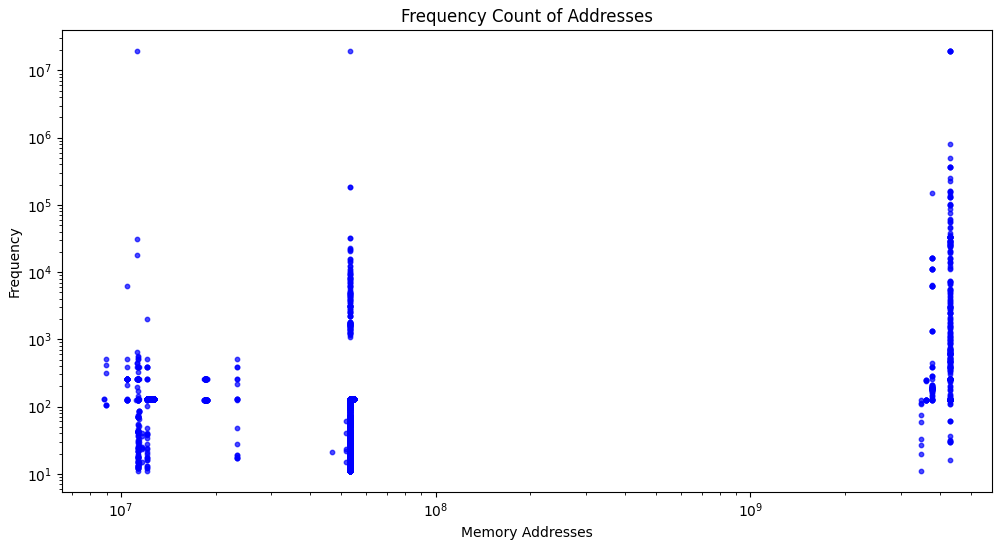}
\caption{Address vs Number of Times Accessed - Split}
\label{fig:freq_plot_split}
\end{figure}

The memory analysis yielded some promising information. As shown in Figure \ref{tab:freq_perc}, the vast majority of addresses were accessed exactly 128 times, which is the number of tokens generated. The decoding phase of the model is only expectedly contains most of these averages. There are only 40 addresses referenced in the prefill phase not used in the decoder phase.  Figure \ref{fig:freq_plot_full} shows the full range of addresses while Figure \ref{fig:freq_plot_split} shows the frequency of addresses in the decoder phase. The different bands of memory are shown here as well, which will be discussed later on. The most accessed memory addresses are stack variables which are used in matrix multiplication operations, which are notably absent in the decoder phase.

\begin{table}[h]
    \centering
    \begin{tabular}{|c|c|c|}
         \hline
         Number of Times an Address is Accessed & Percentage & Amount of Addresses Accessed \\
         \hline
         1 Time & 0.460\% & 1428 \\
         \hline
         128 Times & 98.06\% & 303875 \\
         \hline
         Other & 1.514\% & 4696 \\
         \hline
    \end{tabular}
    \caption{Frequency Percentages}
    \label{tab:freq_perc}
\end{table}

\begin{figure}[h]
\centering
\includegraphics[width=\textwidth, trim={0 0 0 0.71cm}, clip]{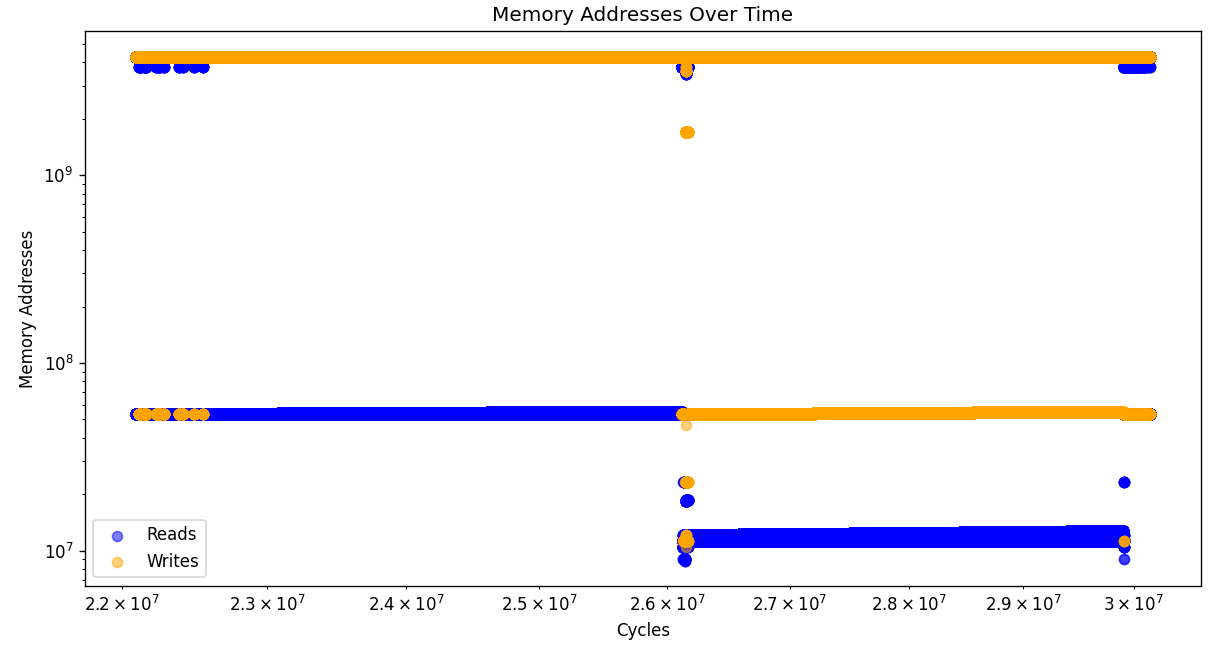}
\caption{Histogram of L1D Stride}
\label{fig:histo_l1d_full}
\end{figure}

\begin{figure}[h]
\centering
\includegraphics[width=\textwidth, trim={0 0 0 0.71cm}, clip]{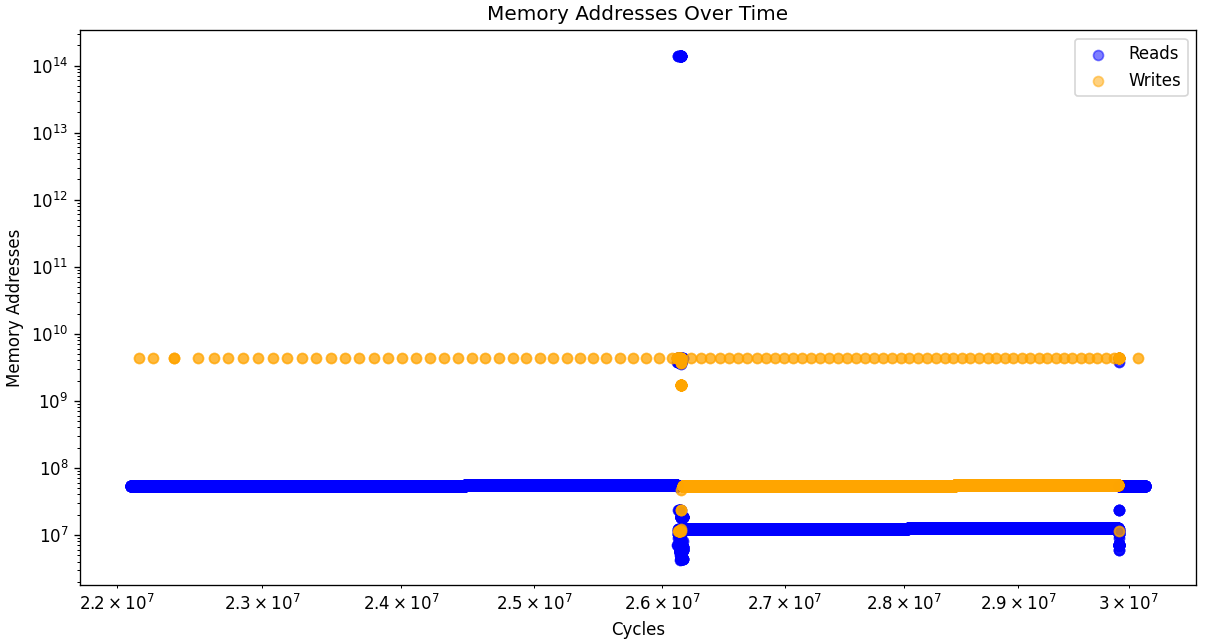}
\caption{Histogram of L2C Stride}
\label{fig:histo_l2c_full}
\end{figure}

\begin{figure}[h]
\centering
\includegraphics[width=\textwidth, trim={0 0 0 0.71cm}, clip]{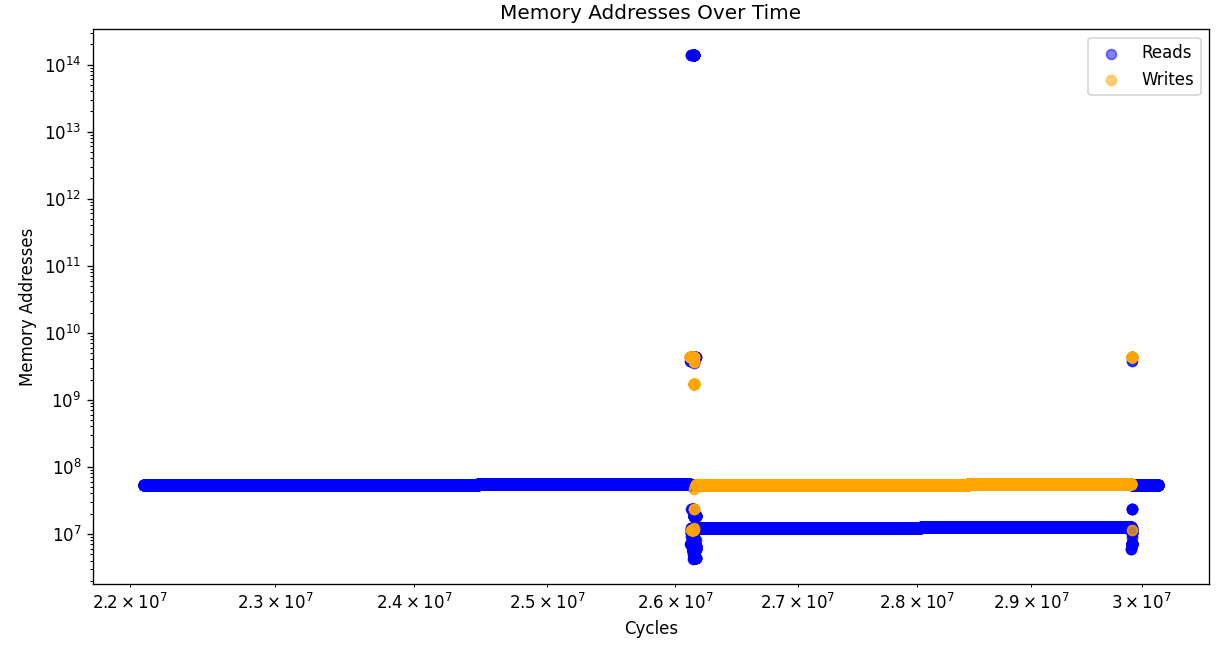}
\caption{Histogram of LLC Stride}
\label{fig:histo_llc_full}
\end{figure}

\begin{figure}[h]
\centering
\includegraphics[width=\textwidth, trim={0 0 0 0.91cm}, clip]{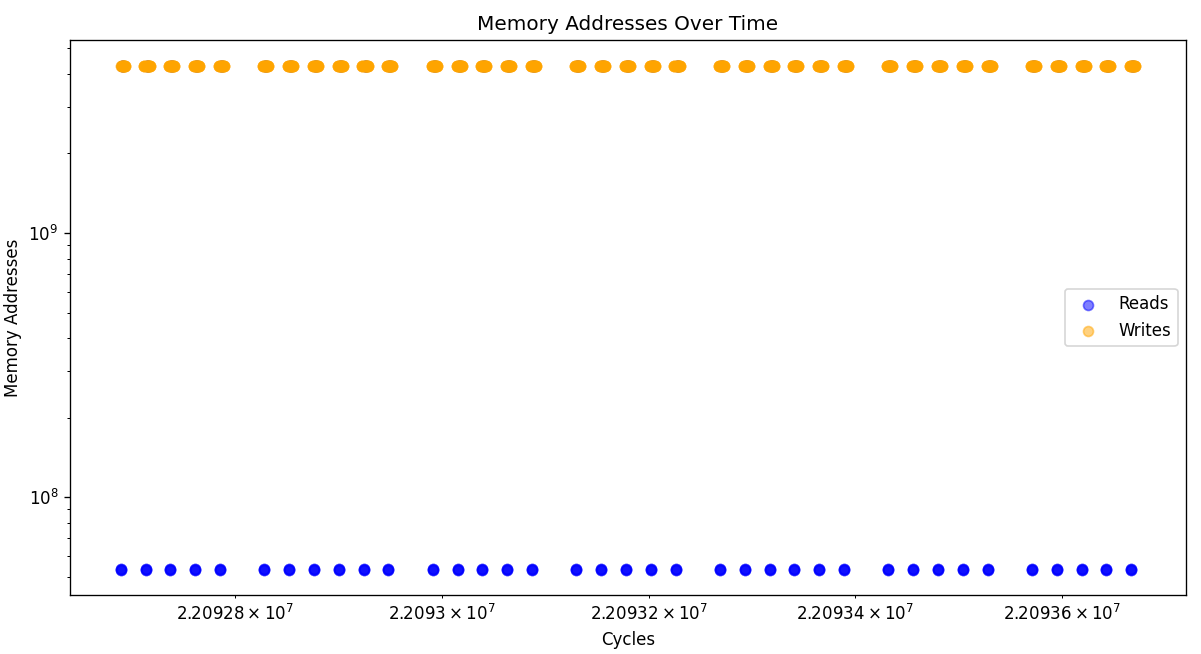}
\caption{First 1k Cycles of L1D Stride}
\label{fig:histo_l1d_1k}
\end{figure}

\begin{figure}[h]
\centering
\includegraphics[width=\textwidth, trim={0 0 0 1.07cm}, clip]{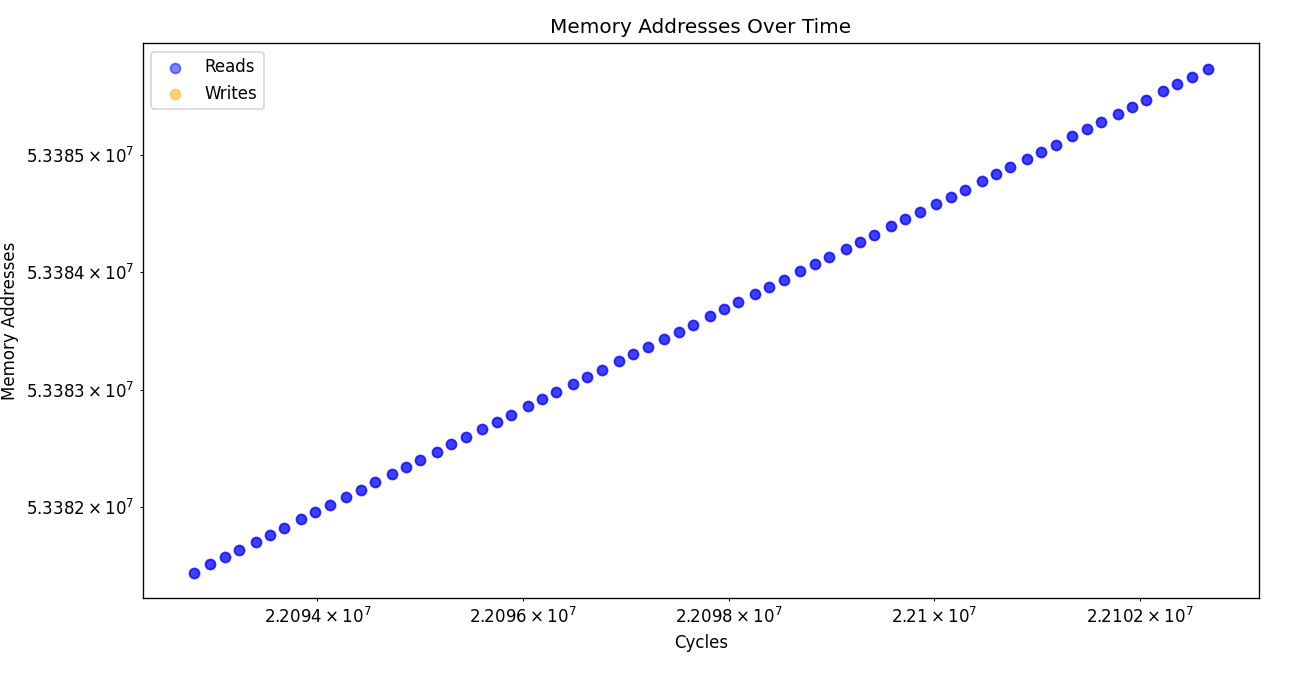}
\caption{First 10k Cycles of L2C Stride}
\label{fig:histo_l2c_10k}
\end{figure}

\begin{figure}[h]
\centering
\includegraphics[width=\textwidth, trim={0 0 0 0.71cm}, clip]{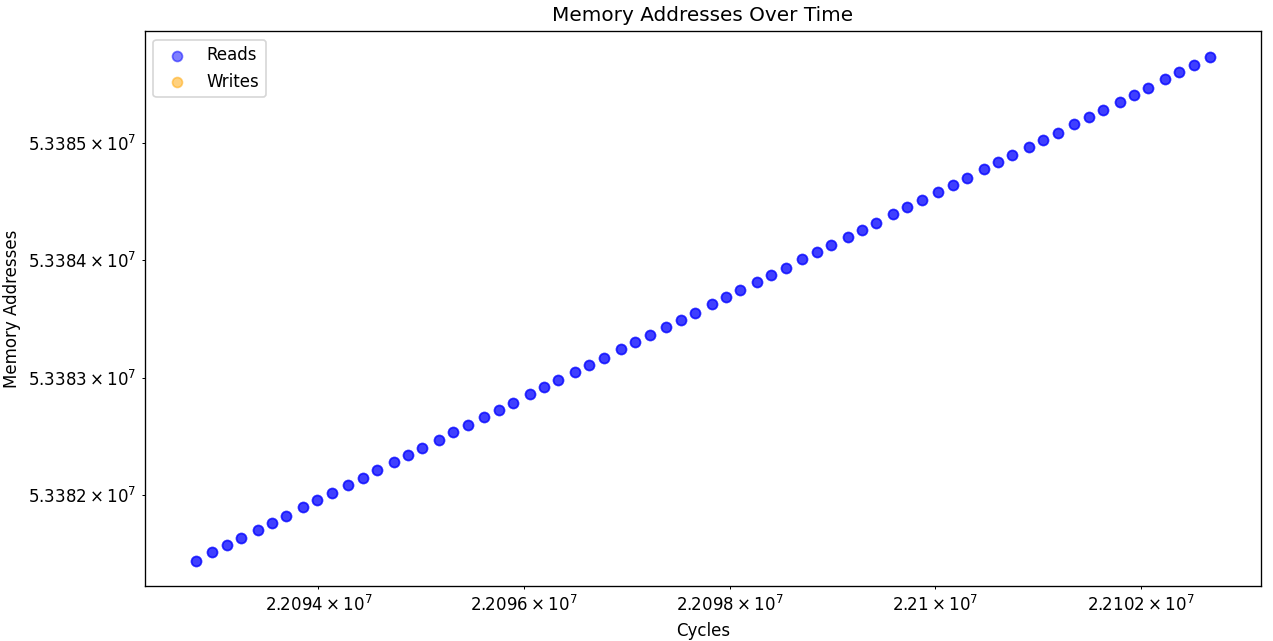}
\caption{First 10k Cycles of LLC Stride}
\label{fig:histo_llc_10k}
\end{figure}

\begin{figure}[h]
\centering
\includegraphics[width=\textwidth, trim={0 0 0 0.8cm}, clip]{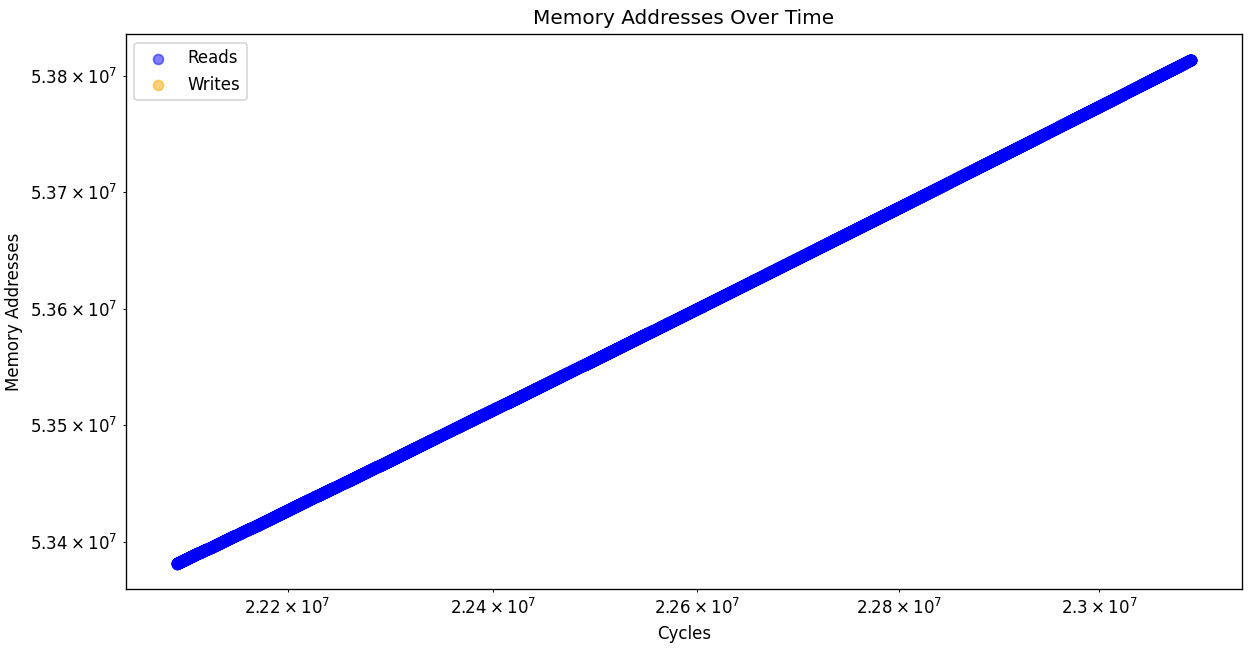}
\caption{First 100k Cycles of LLC Stride}
\label{fig:histo_llc_100k}
\end{figure}

\begin{figure}[h]
\centering
\includegraphics[width=\textwidth, trim={0 0 0 1.06cm}, clip]{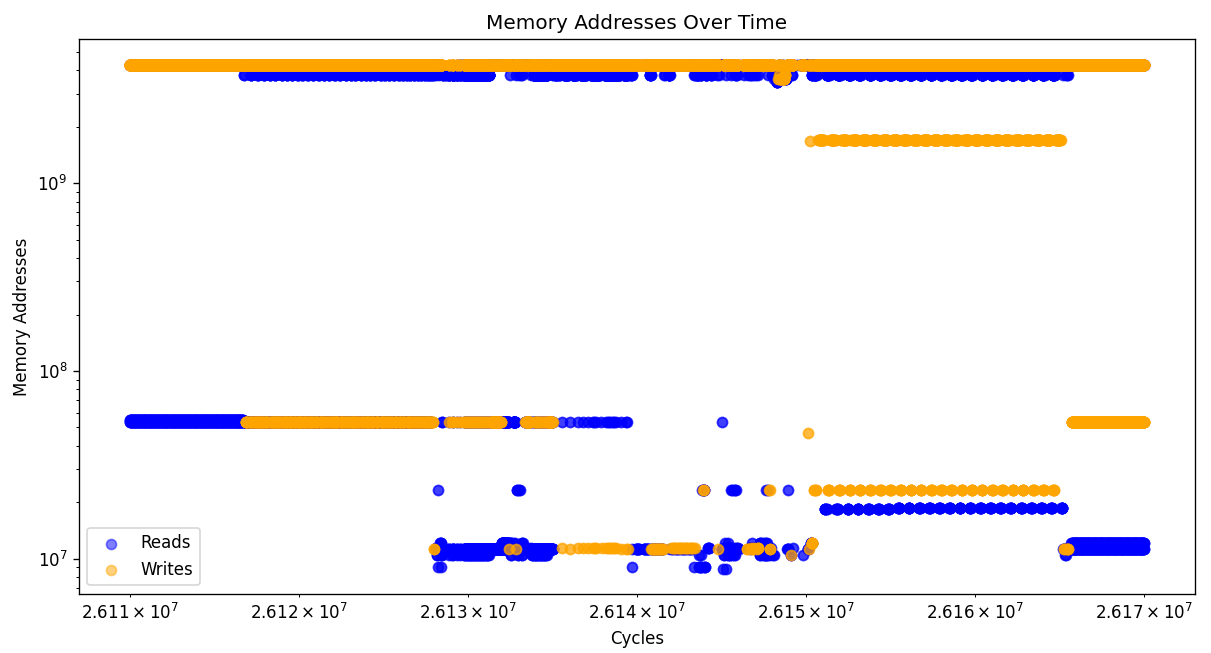}
\caption{L1D Stride Region of Interest}
\label{fig:histo_l1d_roi}
\end{figure}

\begin{figure}[h]
\centering
\includegraphics[width=\textwidth, trim={0 0 0 0.9cm}, clip]{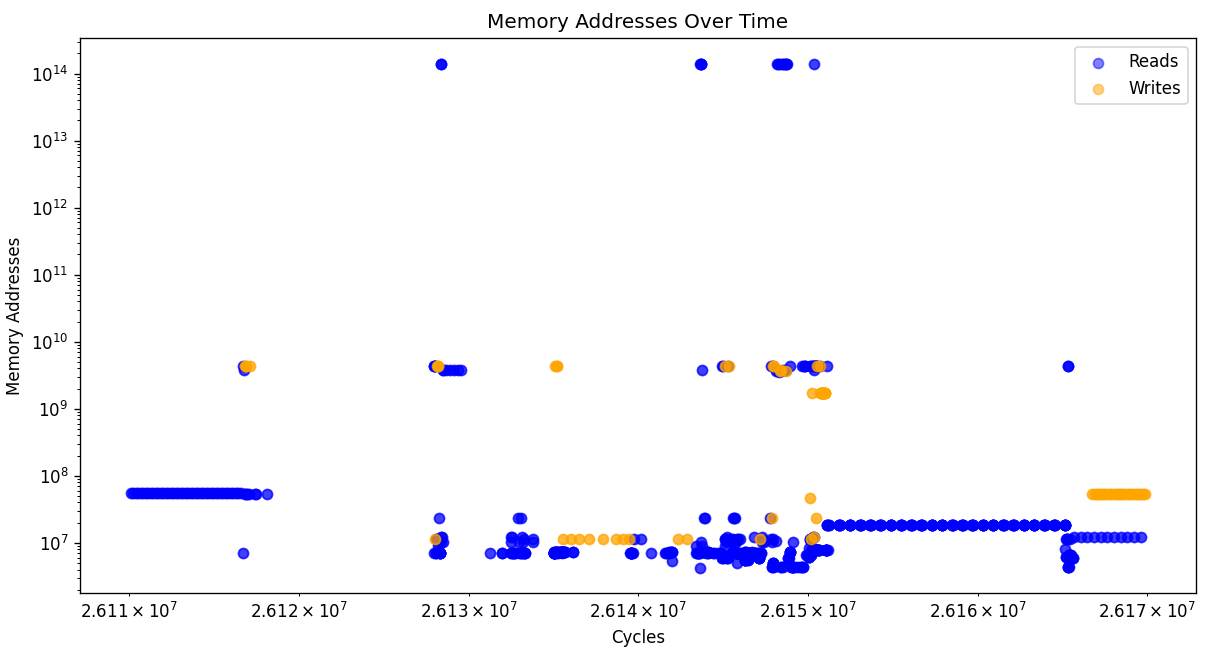}
\caption{L2C Stride Region of Interest}
\label{fig:histo_l2c_roi}
\end{figure}

\begin{figure}[h]
\centering
\includegraphics[width=\textwidth, trim={0 0 0 1.1cm}, clip]{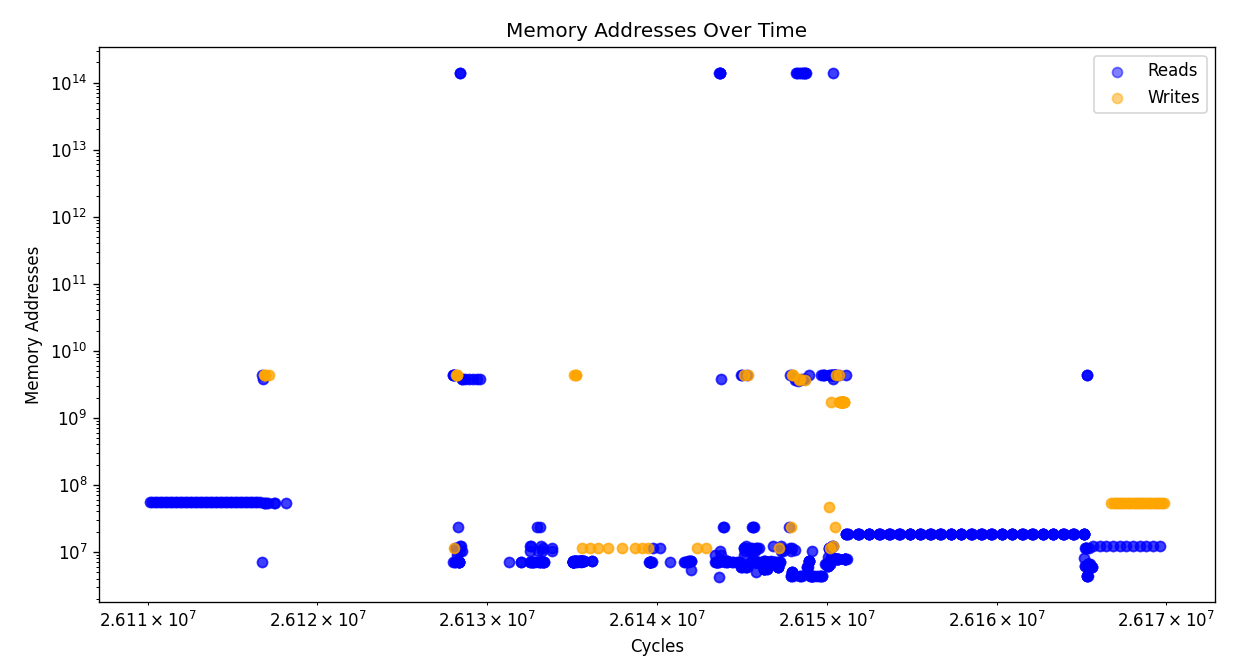}
\caption{LLC Stride Region of Interest}
\label{fig:histo_llc_roi}
\end{figure}

The histograms also provided promising information. Table \ref{tab:histo_stride} shows the cycle numbers and the delta between two cycle numbers for a given address. This address is apart of an array in Llama.cpp that stores token data and is accessed 128 times, making it a good candidate to determine the frequency of a models execution for a given token. This stride was used to create the series of histograms below and capture the memory footprint for the decoding process of a single token. Figures \ref{fig:histo_l1d_full} - \ref{fig:histo_llc_roi} will be discussed in the next section.

\begin{table}[h]
    \centering
    \begin{tabular}{|c|c|}
         \hline
         Cycle Number & Stride \\
         \hline
         13,868,707 & 8,223,982 \\
         \hline
         22,092,689 & 8,058,331 \\
         \hline
         30,151,020 & 8,074,101 \\
         \hline
         38,225,121 & 8,074,533  \\
         \hline
         46,299,654 & 8,039,780  \\
         \hline
         54,339,434 & 8,067,323 \\
         \hline
         62,406,757 & 8,037,652 \\
         \hline
         70,444,409 & 8,058,026 \\
         \hline
         78,502,435 & 8,060,904 \\
         \hline
         86563339 &  \\
         \hline
    \end{tabular}
    \caption{Cycle Stride Table}
    \label{tab:histo_stride}
\end{table}

\chapter{\uppercase {discussion}}

There are promising insights shown here that could greatly increase the performance of the memory-bound decoder phase of an LLM. The simulations show that Bingo, Berti, and DRRIP perform best, but to put reasons behind the simulations of the performance, we must first analyze the memory access frequency count and histograms. With the context provided from the data and knowledge of how the prefetching and replacement policies work, we will understand why those implementations performed best.

Per Figure \ref{tab:freq_perc}, 98.06\% addresses are accessed exactly 128 times. Given that many of these addresses are accessed once per token and the delta is 8 million cycles, it is likely that many of these addresses will be evicted once. Of the 303,875 addresses in this category, over half of them are related to token data. Recall that the address used to calculate the strides in Figure \ref{tab:histo_stride}, 0x32e8910, was apart of an array storing this token data. This array has 151,936 elements at 24 bytes each, taking up 3,646,464 bytes of memory. The address accessed most frequently of this array was 0x32d6544, accessed 19,445,912 times. This is likely the first element of the array and the address would be used as an offset for all array accesses. When adding the size of the array to the first element, we find that the token data array reaches to 0x3650944. In decimal, that address is $5.69 \times 10^7$, corresponding to the middle right memory band in Figure \ref{fig:freq_plot_full}. With this in mind, any address in that band is accessing token data.

With the knowledge of the memory address bands, we can put meaning to the histograms. Figure \ref{fig:histo_l1d_full} and \ref{fig:histo_llc_full} show that most stack-allocated addresses stay within the L1D or the L2C and almost of the data coming from RAM in a token delta will be token data. The lower band in Figure \ref{fig:histo_llc_full} ranging from $2.6 \times 10^7$ to $3.0 \times 10^7$ are an array of logits accessed by a context variable, named ctx in Llama.cpp. This occurs Llama's set\_logits function where token data is constructed using the logits.

Figure \ref{fig:histo_l1d_1k} shows a closer look at the delta patterns in the memory accesses. Importantly, there are two bands of spatial locality that must be kept track of. This should not be much of an issue though, since the changes within the bands are incremental and remain within page boundaries. The L2C and LLC are much more consistent with its stride patterns Figure \ref{fig:histo_l2c_10k} and \ref{fig:histo_llc_10k} are identical. \ref{fig:histo_llc_100k} continues the linear pattern shown in the ten thousand spans.

Concerning when writes occur, the range discussed earlier in Figure \ref{fig:histo_llc_full} shows that the logits are used to write to the token data. This is a consistent pattern across decoding phases. A notable area of memory occurs at the beginning of this range, from $2.611 \times 10^7$ to $2.617 \times 10^7$. Figures \ref{fig:histo_l1d_roi}, \ref{fig:histo_l2c_roi}, and \ref{fig:histo_llc_roi} show this region of interest in the different areas. These graphs demonstrate a complicated and fairly unpredictable stride pattern compared to other areas, which could be difficult for a prefetcher to handle.

These histograms show a wide working set with a stride pattern that stays mostly consistent but has areas that are unpredictable. Stride prefetchers utilizing a page by page delta will be able to differentiate between the different bands, and \ref{tab:pref_l1d} supports this idea with Bingo and IPCP having lower miss rates than Next Line. The L2C histograms matching the LLC histograms show that the L2C is unable to support the size of the working set, while the LLC is able to keep the vocabulary and token data within its memory. It is important to consider that this will not hold for larger models with higher parameter counts, and for situations where more tokens are generated. There is still a lack of temporal locality, with length re-reference intervals that DRRIP or SHiP will need to make use of.

The simpler stride pattern in the L2C provides an opportunity for a prefetcher such as Berti or Next Line to leverage their abilities and significantly reduce the amount of hits in the cache. A simple prefetcher with low latency may suffice for the L1D cache, but the L2C should seek a prefetcher that provides a low miss percentage without introducing much prefetch traffic. The L1D prefetcher will pollute the L2Cs space with its own prefetches, but as the number of accesses increase, reducing the miss percentage of the L1D becomes more important.
%
%



\chapter{\uppercase {Conclusion}}

\section{Future Work}

There is much more that could be researched in this area, especially in the context of a multithreaded environment. As LLMs benefit significantly from parallelization, it is important that prefetching and replacement policies perform well in multi-core environments. Since CPUs should be general purpose, software based prefetching should be looked into, alongside investigating the impacts of large page sizes to reduce the jumps in physical memory.
Additional parameters such as batch size and context size could be tweaked to investigate performance implications. More complex models or higher token counts could be looked at to see how this impacts the LLC, since the miss rates should drastically increase as the working set exceeds the 4MB capacity of the cache.
Finally, work can be done to use Intel Pin to annotate variables in the source code to provide more certainty in mapping variables to addresses to obtain a clearer picture of how the model executes in memory.

\section{Conclusion}

We aimed to characterize the memory access patterns of a large language model in a CPU environment to discover the impacts of these patterns, and the insights that could be found from this investigation. CPUs offer a higher memory capacity and have other benefits that make them an attractive candidate for running inference if their execution time can be decreased. With the memory demands from increased model sizes, the KV Cache used to decrease execution time at the cost of memory, and the majority of execution time spent in the memory-bound decoder phase, memory efficiency is crucial to the performance of an LLM.

We find that the majority of memory accesses in an LLM are due to accessing the vocabulary of the model, updating its weights and logits, over a long period of cycles. A large working set combined with a long re-reference interval shows that a cache replacement policy focusing on this could greatly improve performance. The predictable stride of the L2C cache and its simulation results show that the L2C runs out of space, and even a simple prefetcher would improve performance. Closer to the CPU, the memory access patterns become more complex, swapping through the two bands of memory. There are significant improvements that could be made to the memory efficiency of the cache system when it comes to LLMs.


\let\oldbibitem\bibitem
\renewcommand{\bibitem}{\setlength{\itemsep}{0pt}\oldbibitem}
\bibliographystyle{ieeetr}

\phantomsection
\addcontentsline{toc}{chapter}{REFERENCES}

\renewcommand{\bibname}{{\normalsize\rm REFERENCES}}

\bibliography{my_ref}

\begin{thebibliography}{10}

\bibitem{cache-deep-learning}
Z.~Shi, X.~Huang, A.~Jain, and C.~Lin, ``Applying deep learning to the cache
  replacement problem,'' in {\em Proceedings of the 52nd Annual IEEE/ACM
  International Symposium on Microarchitecture}, MICRO '52, (New York, NY,
  USA), p.~413–425, Association for Computing Machinery, 2019.

\bibitem{zeroshot}
T.~B. Brown, B.~Mann, N.~Ryder, M.~Subbiah, J.~Kaplan, P.~Dhariwal,
  A.~Neelakantan, P.~Shyam, G.~Sastry, A.~Askell, S.~Agarwal, A.~Herbert-Voss,
  G.~Krueger, T.~Henighan, R.~Child, A.~Ramesh, D.~M. Ziegler, J.~Wu,
  C.~Winter, C.~Hesse, M.~Chen, E.~Sigler, M.~Litwin, S.~Gray, B.~Chess,
  J.~Clark, C.~Berner, S.~McCandlish, A.~Radford, I.~Sutskever, and D.~Amodei,
  ``Language models are few-shot learners,'' 2020.

\bibitem{bert}
J.~Devlin, M.-W. Chang, K.~Lee, and K.~N. Toutanova, ``Bert: Pre-training of
  deep bidirectional transformers for language understanding,'' 2018.

\bibitem{openai2024gpt4technicalreport}
{OpenAI (2023)}, ``Gpt-4 technical report,'' 2024.

\bibitem{qwen}
J.~Bai, S.~Bai, Y.~Chu, Z.~Cui, K.~Dang, X.~Deng, Y.~Fan, W.~Ge, Y.~Han,
  F.~Huang, B.~Hui, L.~Ji, M.~Li, J.~Lin, R.~Lin, D.~Liu, G.~Liu, C.~Lu, K.~Lu,
  J.~Ma, R.~Men, X.~Ren, X.~Ren, C.~Tan, S.~Tan, J.~Tu, P.~Wang, S.~Wang,
  W.~Wang, S.~Wu, B.~Xu, J.~Xu, A.~Yang, H.~Yang, J.~Yang, S.~Yang, Y.~Yao,
  B.~Yu, H.~Yuan, Z.~Yuan, J.~Zhang, X.~Zhang, Y.~Zhang, Z.~Zhang, C.~Zhou,
  J.~Zhou, X.~Zhou, and T.~Zhu, ``Qwen technical report,'' 2023.

\bibitem{llm-cpu-perf}
S.~Na, G.~Jeong, B.~H. Ahn, J.~Young, T.~Krishna, and H.~Kim, ``{ Understanding
  Performance Implications of LLM Inference on CPUs },'' in {\em 2024 IEEE
  International Symposium on Workload Characterization (IISWC)}, (Los Alamitos,
  CA, USA), pp.~169--180, IEEE Computer Society, Sept. 2024.

\bibitem{nomad}
T.~Zhang, J.~W. Yi, B.~Yao, Z.~Xu, and A.~Shrivastava, ``Nomad-attention:
  Efficient llm inference on cpus through multiply-add-free attention,'' 2024.

\bibitem{cpu-gpu-orch}
K.~Kamahori, Y.~Gu, K.~Zhu, and B.~Kasikci, ``Fiddler: Cpu-gpu orchestration
  for fast inference of mixture-of-experts models,'' 2024.

\bibitem{memorywall}
A.~Gholami, Z.~Yao, S.~Kim, C.~Hooper, M.~W. Mahoney, and K.~Keutzer, ``Ai and
  memory wall,'' 2024.

\bibitem{sgx}
V.~Costan and S.~Devadas, ``Intel {SGX} explained.'' Cryptology {ePrint}
  Archive, Paper 2016/086, 2016.

\bibitem{tinyml-review}
P.~P. Ray, ``A review on tinyml: State-of-the-art and prospects,'' {\em Journal
  of King Saud University - Computer and Information Sciences}, vol.~34, no.~4,
  pp.~1595--1623, 2022.

\bibitem{power-consumption}
D.~Li, X.~Chen, M.~Becchi, and Z.~Zong, ``Evaluating the energy efficiency of
  deep convolutional neural networks on cpus and gpus,'' in {\em 2016 IEEE
  International Conferences on Big Data and Cloud Computing (BDCloud), Social
  Computing and Networking (SocialCom), Sustainable Computing and
  Communications (SustainCom) (BDCloud-SocialCom-SustainCom)}, pp.~477--484,
  2016.

\bibitem{tiny-ml}
Y.~Abadade, A.~Temouden, H.~Bamoumen, N.~Benamar, Y.~Chtouki, and A.~S. Hafid,
  ``A comprehensive survey on tinyml,'' {\em IEEE Access}, vol.~11,
  pp.~96892--96922, 2023.

\bibitem{cpu-vs-gpu-dl}
E.~BUBER and B.~DIRI, ``Performance analysis and cpu vs gpu comparison for deep
  learning,'' in {\em 2018 6th International Conference on Control Engineering
  \& Information Technology (CEIT)}, pp.~1--6, 2018.

\bibitem{tiny-ml-benchmark}
C.~R. Banbury, V.~J. Reddi, M.~Lam, W.~Fu, A.~Fazel, J.~Holleman, X.~Huang,
  R.~Hurtado, D.~Kanter, A.~Lokhmotov, D.~Patterson, D.~Pau, J.~sun Seo,
  J.~Sieracki, U.~Thakker, M.~Verhelst, and P.~Yadav, ``Benchmarking tinyml
  systems: Challenges and direction,'' 2021.

\bibitem{temporal}
H.~Wu, K.~Nathella, J.~Pusdesris, D.~Sunwoo, A.~Jain, and C.~Lin, ``Temporal
  prefetching without the off-chip metadata,'' in {\em Proceedings of the 52nd
  Annual IEEE/ACM International Symposium on Microarchitecture}, MICRO '52,
  (New York, NY, USA), p.~996–1008, Association for Computing Machinery,
  2019.

\bibitem{stride}
J.~W. Fu, J.~H. Patel, and B.~L. Janssens, ``Stride directed prefetching in
  scalar processors,'' {\em ACM SIGMICRO Newsletter}, vol.~23, no.~1-2,
  pp.~102--110, 1992.

\bibitem{berti}
A.~Ros, ``Berti: A per-page best-request-time delta prefetcher,'' 2019.

\bibitem{ipcp}
S.~Pakalapati and B.~Panda, ``Bouquet of instruction pointers: Instruction
  pointer classifier-based spatial hardware prefetching,'' in {\em 2020
  ACM/IEEE 47th Annual International Symposium on Computer Architecture
  (ISCA)}, pp.~118--131, 2020.

\bibitem{bingo}
M.~Bakhshalipour, M.~Shakerinava, P.~Lotfi-Kamran, and H.~Sarbazi-Azad,
  ``Accurately and maximally prefetching spatial data access patterns with
  bingo,'' 2019.

\bibitem{SPP}
J.~Kim, S.~H. Pugsley, P.~V. Gratz, A.~N. Reddy, C.~Wilkerson, and Z.~Chishti,
  ``Path confidence based lookahead prefetching,'' in {\em 2016 49th Annual
  IEEE/ACM International Symposium on Microarchitecture (MICRO)}, pp.~1--12,
  2016.

\bibitem{drrip}
A.~Jaleel, K.~B. Theobald, S.~C. Steely, and J.~Emer, ``High performance cache
  replacement using re-reference interval prediction (rrip),'' in {\em
  Proceedings of the 37th Annual International Symposium on Computer
  Architecture}, ISCA '10, (New York, NY, USA), p.~60–71, Association for
  Computing Machinery, 2010.

\bibitem{SHip}
C.-J. Wu, A.~Jaleel, W.~Hasenplaugh, M.~Martonosi, S.~C. Steely, and J.~Emer,
  ``Ship: signature-based hit predictor for high performance caching,'' in {\em
  Proceedings of the 44th Annual IEEE/ACM International Symposium on
  Microarchitecture}, MICRO-44, (New York, NY, USA), p.~430–441, Association
  for Computing Machinery, 2011.

\bibitem{cache-bypass}
S.~Mittal, ``A survey of cache bypassing techniques,'' {\em Journal of Low
  Power Electronics and Applications}, vol.~6, no.~2, 2016.

\bibitem{champsim}
N.~Gober, G.~Chacon, L.~Wang, P.~V. Gratz, D.~A. Jimenez, E.~Teran, S.~Pugsley,
  and J.~Kim, ``The championship simulator: Architectural simulation for
  education and competition,'' 2022.

\end{thebibliography}

\include{data/appendices}

\end{document}